\documentclass[10pt,twocolumn,letterpaper]{article}

\usepackage[pagenumbers]{cvpr} 

\usepackage[dvipsnames]{xcolor}

\definecolor{cvprblue}{rgb}{0.21,0.49,0.74}
\definecolor{gold}{rgb}{1.0, 0.84, 0.0}
\definecolor{silver}{rgb}{0.75, 0.75, 0.75}

\usepackage[pagebackref,breaklinks,colorlinks,citecolor=cvprblue]{hyperref}

\usepackage{soul}

\usepackage{graphicx}
\usepackage{amsmath}
\usepackage{amssymb}
\usepackage{booktabs}
\usepackage{array}
\usepackage{multirow}
\usepackage{bbding}

\usepackage{caption}
\usepackage{lipsum}
\usepackage{subcaption}
\newcommand{\mycirc}[1][black]{\Large\textcolor{#1}{\ensuremath\bullet}}
\usepackage{appendix}

\usepackage{algorithm}
\usepackage{algpseudocode}
\usepackage{hyperref}

\def\paperID{5179} 
\def\confName{CVPR}
\def\confYear{2024}

\title{MVDD: Multi-View Depth Diffusion Models}

\author{Zhen Wang$^{1,2*}$ \qquad Qiangeng Xu$^{1}$ \qquad Feitong Tan$^{1}$ \qquad Menglei Chai$^{1}$ \qquad Shichen Liu$^{1}$ \\ \qquad  Rohit Pandey$^{1}$ \qquad Sean Fanello$^{1}$ \qquad Achuta Kadambi$^{1,2}$ \qquad Yinda Zhang$^{1}$ \\
	\hspace{0mm}$^1$Google \hspace{18mm} 
	$^2$University of California, Los Angeles\hspace{18mm}
}

\begin{document}

\newcommand{\ak}[1]{\textcolor{magenta}{ak: #1}}
\newcommand{\charlie}[1]{\textcolor{cyan}{charlie: #1}}

\twocolumn[{%
	\renewcommand\twocolumn[1][]{#1}%
	\maketitle
	\captionsetup{type=figure}
	\vspace{-2mm}
	\centering{
		\includegraphics[width=0.97\textwidth]{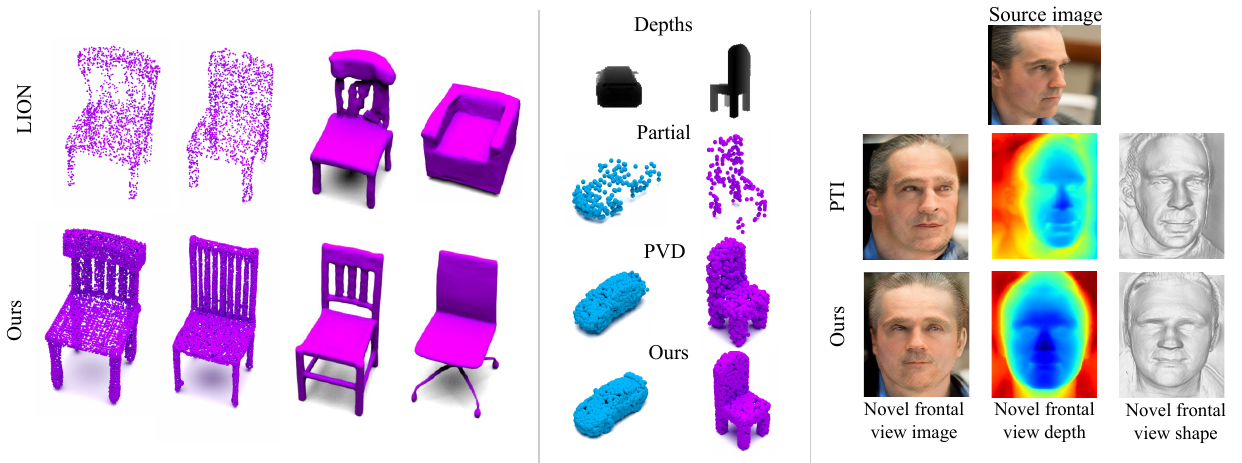}\vspace{-4mm}
		\captionsetup[sub]{font=normal,labelfont={bf,sf}}
		\hfill\subcaptionbox{ Shape generation.}[0.37\linewidth]
		
		\hfill\subcaptionbox{Shape completion.}[0.17\linewidth]
		\hfill\subcaptionbox{Shape prior for 3D GAN inversion.}[0.35\linewidth]
		\hfill \vspace{-1mm}
		\hfill\caption{Our proposed MVDD is versatile and can be utilized in various applications: (a) 3D shape generation: our model generates high-quality 3D shape with approximately 10X more points than diffusion-based point cloud generative models e.g., LION~\cite{vahdat2022lion} and PVD~\cite{zhou20213d}  and contains diverse and fine-grained details. (b) Shape completion: we showcase shape completion results from partial inputs, highlighting the higher fidelity compared to PVD~\cite{zhou20213d}.  (c) Our model can serve as a powerful shape prior for downstream tasks such as 3D GAN inversion~\cite{chan2022efficient, roich2022pivotal}.}
		\label{fig:teaser}
	}
	\hfill \vspace{0mm}
}]
\newcommand{\DepthFusion}{denoising depth fusion}
\newcommand{\DepthFusionCap}{Denoising Depth Fusion}

\let\thefootnote\relax\footnotetext{*Work done while the author was an intern at Google. See our web page at \url{https://mvdepth.github.io/}}

\begin{abstract}
Denoising diffusion models have demonstrated outstanding results in 2D image generation, yet it remains a challenge to replicate its success in 3D shape generation. In this paper, we propose leveraging multi-view depth, which represents complex 3D shapes in a 2D data format that is easy to denoise. We pair this representation with a diffusion model, MVDD, that is capable of generating high-quality dense point clouds with 20K+ points with fine-grained details. To enforce 3D consistency in multi-view depth, we introduce an epipolar line segment attention that conditions the denoising step for a view on its neighboring views. Additionally, a depth fusion module is incorporated into diffusion steps to further ensure the alignment of depth maps. When augmented with surface reconstruction, MVDD can also produce high-quality 3D meshes. Furthermore, MVDD stands out in other tasks such as depth completion, and can serve as a 3D prior, significantly boosting many downstream tasks, such as GAN inversion. State-of-the-art results from extensive experiments demonstrate MVDD's excellent ability in 3D shape generation, depth completion, and its potential as a 3D prior for downstream tasks.

\end{abstract}

\section{Introduction}
\label{sec:intro}

3D shape generative models have made remarkable progress in the wave of AI-Generated Content. A powerful 3D generative model is expected to possess the following attributes: (i) \textit{Scalability}. The model should be able to create objects with fine-grained details; (ii) \textit{Faithfulness}. The generated 3D shapes should exhibit high fidelity and resemble the objects in the dataset; and (iii) \textit{Versatility}. The model can be plugged in as a 3D prior in various downstream 3D tasks through easy adaptation. Selecting suitable probabilistic models becomes the key factor in achieving these criteria. Among popular generative methods such as GANs \cite{goodfellow2020generative,lan2023e3dge}, VAEs \cite{kingma2013auto}, and normalizing flows \cite{papamakarios2021normalizing}, denoising diffusion models \cite{ho2020denoising,song2020denoising} explicitly model the data distribution; therefore, they are able to faithfully generate images that reflect content diversity.

It is also important to choose suitable 3D representations for shape generation. While delivering high geometric quality and infinite resolution, implicit function-based models \cite{mittal2022autosdf,nam20223d,chen2019learning,chou2023diffusion,xu2019disn} tend to be computationally expensive. This is due to the fact that the number of inferences increases cubically with the resolution and the time-consuming post-process, e.g., marching cubes. On the other hand, studies~\cite{luo2021diffusion,zhou20213d,vahdat2022lion} learn diffusion models on a point cloud by adding noise and denoising either directly on point positions or their latent embeddings. Due to the irregular data format of the point set, it requires over 10,000 epochs for these diffusion models to converge on a single ShapeNet \cite{chang2015shapenet} category, while the number of points that can be generated by these models typically hovers around 2048. 

In this work, we investigate a multi-view depth representation and propose a novel diffusion model, namely MVDD, which generates 3D consistent multi-view depth maps for 3D shape generation. The benefits of using the multi-view depth representation with diffusion models come in three folds: 1) The representation is naturally supported by diffusion models. The 2D data format conveniently allows the direct adoption of powerful 2D diffusion architectures  \cite{radford2021learning,deepfloyd}; 2) Multi-view depth registers complex 3D surfaces onto 2D grids, essentially reducing the dimensionality of the 3D generation space to 2D. As a result, the generated 2D depth map can have higher resolution than volumetric implicit representations \cite{mittal2022autosdf} and produce dense point clouds with a much higher point count; 3) Depth is a widely used representation; therefore, it is easy to use it as a 3D prior to support downstream applications.

While bearing this many advantages, one key challenge of using multi-view depths for 3D shape generation is cross-view consistency. Even with a well-trained diffusion model that learns the depth distribution from 3D consistent depth maps, the generated multi-view depth maps are not guaranteed to be consistent after ancestral sampling \cite{luo2021diffusion}. To tackle this challenge, our proposed MVDD conditions diffusion steps for each view on neighboring views, allowing different views to exchange information. This is realized by a novel epipolar ``line segment'' attention, which benefits from epipolar geometry. 
Differing from full attention \cite{shi2023mvdream} and epipolar attention \cite{liu2023syncdreamer}, our epipolar ``line segment'' attention leverages the depth estimation in our diffusion process. Therefore, it only attends to features at the most relevant locations, making it both effective and efficient. However, even with relatively consistent multi-view maps, back-projected 3D points from each depth map are still not guaranteed to be perfectly aligned, resulting in ``double layers'' in the 3D shapes (see \cref{fig:depth_fusion}(c)). To address this issue, MVDD incorporates depth fusion in denoising steps to explicitly align depth from multiple views.

Empowered by these modules, MVDD can generate high-quality 3D shapes, faithfully conduct depth completion, and distill 3D prior knowledge for downstream tasks. We summarize our contributions as follows:
\begin{itemize}
    
    \item {To the best of our knowledge, we propose the first multi-view depth representation in the generative setting with a diffusion model MVDD. The representation reduces the dimension of the generation space and avoid unstructured data formats such as point set. Therefore, it is more scalable and suitable for diffusion frameworks and is easier to converge.}
    
    \item {We also propose an epipolar ``line segment'' attention and \DepthFusion~ that could effectively enforce 3D consistency for multi-view depth maps.}
    
    
    \item {Through extensive experiments, we show the flexibility and versatility of MVDD in various tasks such as 3D shape generation and shape completion. Our method outperforms compared methods in both shape generation and shape completion by substantial margins.}
    
\end{itemize}

\begin{figure*}[t]
  \centering
   \includegraphics[width=\linewidth]{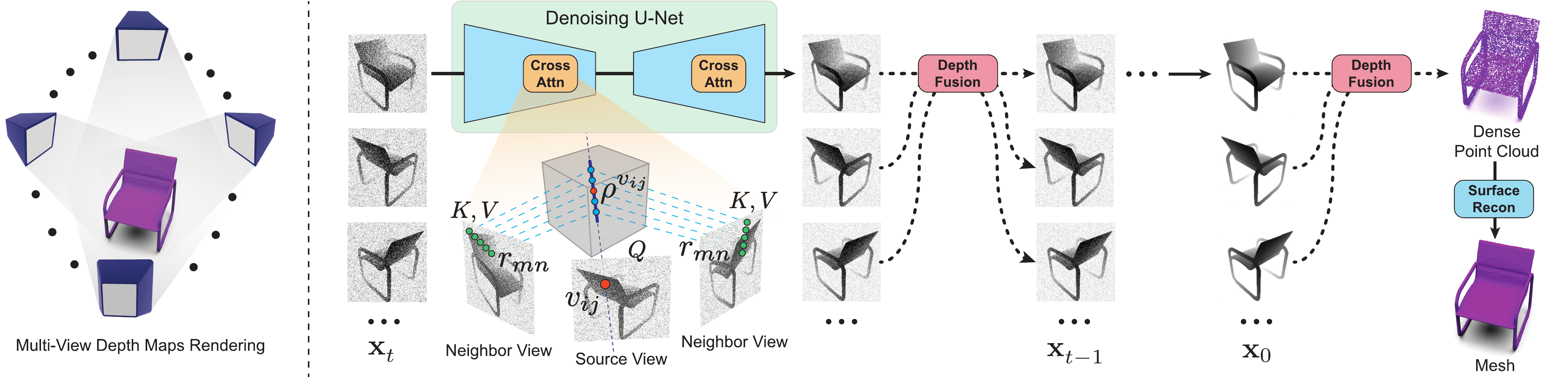}
   \caption{Our method collects ground truth from multi-view rendered depth maps (left). Starting with multiple 2D maps with randomly sampled noise, MVDD generates diverse multi-view depth through an iterative denoising diffusion process (right). To enforce multi-view 3D consistency, MVDD denoises each depth map with an efficient epipolar ``line segment'' attention (\cref{sec:epipolar}). Specifically, by leveraging the denoised value from the current step, MVDD only needs to attend to features on a line segment centered around the back-projected depth (the red dot), rather than the entire epipolar line. To further align the denoised multi-view depths, depth fusion (\cref{sec:fusion}) is incorporated after the U-Net in a denoising step. The final multi-view depth can be fused together to obtain a high-quality dense point cloud, which can then be reconstructed into high quality 3D meshes with fine-grained details.}
   \label{fig:pipeline}
\end{figure*}
\section{Related Work}
\label{sec:related_work}

\subsection{3D Shape Generative Models}
Representations such as implicit functions, voxels, point clouds, and tetrahedron meshes have been used for 3D shape generation in previous studies.

Implicit-based models, such as AutoSDF~\cite{mittal2022autosdf}, infer SDF from feature volumes. Since the computation for volumes grows cubically with resolution, the volume resolution is often limited. Voxel-based models, such as Vox-Diff \cite{zhou20213d}, face the same challenge. Other implicit-based models, such as 3D-LDM~\cite{nam20223d}, IM-GAN~\cite{chen2019learning}, and Diffusion-sdf~\cite{chou2023diffusion}, generate latent codes and use auto-encoders to infer SDFs. The latent solution helps avoid the limitation of resolution but is prone to generate overly smoothed shapes. When combined with tetrahedron mesh, implicit methods \cite{gao2022get3d,liu2023meshdiffusion} are able to generate  compact implicit fields and achieve high-quality shape generation. However, unlike multi-view depth, it is non-trivial for them to serve as a 3D prior in downstream tasks that do not use tetrahedron grids.

Point cloud-based methods avoid modeling empty space inherently. Previous explorations include SetVAE~\cite{kim2021setvae} and VG-VAE~\cite{anvekar2022vg}, which adopt VAEs for point latent sampling. GAN-based models \cite{valsesia2018learning,shu20193d} employ adversarial loss to generate  point clouds. Flow-based models \cite{klokov2020discrete,yang2019pointflow} use affine coupling layers to model point distributions. To enhance generation diversity, some studies leverage diffusion~\cite{ho2020denoising,song2020denoising} to generate 3D point cloud distributions. ShapeGF~\cite{cai2020learning} applies the score-matching gradient flow to move the point set. DPM~\cite{luo2021diffusion} and PVD~\cite{zhou20213d} denoise Gaussian noise on point locations. LION \cite{vahdat2022lion} encodes the point set into latents and then conducts latent diffusion. Although these models excel in producing diverse shapes, the denoising scheme on unstructured point cloud data limits the number of points that can be generated. Our proposed model leverages multi-view depth representation, which can generate high-resolution point clouds, leading to 3D shapes with fine details.

\subsection{Multi-View Diffusion Models}
The infamous Janus problem~\cite{poole2022dreamfusion, melas2023realfusion} and 3D inconsistency have plagued SDS-based \cite{poole2022dreamfusion} 3D content generation. MVDream~\cite{shi2023mvdream} connects rendered images from different views via a 3D self-attention fashion to constrain multi-view consistency in the generated multi-view images. SyncDreamer~\cite{liu2023syncdreamer} builds a cost volume that correlates the corresponding features across different views to synchronize the intermediate states of all the generated images at each step of the reverse process. EfficientDreamer~\cite{zhao2023efficientdreamer} and TextMesh~\cite{tsalicoglou2023textmesh} concatenate canonical views either channel-wise or spatially into the diffusion models to enhance 3D consistency. SweetDreamer~\cite{li2023sweetdreamer} proposes aligned geometry priors by fine-tuning the 2D diffusion models to be viewpoint-aware and to produce view-specific coordinate maps. Our method differs from them in that we generate multi-view depth maps, instead of RGB images, and thus propose an efficient epipolar line segment attention tailored for depth maps to enforce 3D consistency.
\newcommand{\NumNeighbors}{R}
\newcommand{\SourcePixel}{v_{ij}}
\newcommand{\SourcePixelDepth}{\mathbf{x}^{\SourcePixel}_t}
\newcommand{\NeighborPixel}{r_{mn}}
\newcommand{\NeighborPixelDepth}{\mathbf{x}^{\NeighborPixel}_t}
\newcommand{\threeDpoint}{\rho}
\newcommand{\BackPixel}{v_{\tilde{i}\tilde{j}}}
\newcommand{\PixelThreshHold}{\psi_{\max }}

\section{Method}
In this section, we introduce our Multi-View Depth Diffusion Models (MVDD). 
We first provide an overview of MVDD in ~\cref{sec:mvdd}, a model that aims to produce multi-view depth. After that, we illustrate how multi-view consistency is enforced among different views of depth maps in our model by using epipolar ``line segment'' attention (\cref{sec:epipolar}) and \DepthFusion~(\cref{sec:fusion}). Finally, we introduce the training objectives in~\cref{sec:training} and  implementation details in ~\cref{sec:implementation}. 
\subsection{Multi-View Depth Diffusion}
\label{sec:mvdd}
Our method represents a 3D shape $\mathcal{X}$ by using its multi-view depth maps $\mathbf{x} \in \mathbb{R}^{N \times H \times W} = \{\mathbf{x}^v | v = 1,2,...,N \}$, where $v$ is the index of the view, $N$ is the total number of views, and $H$ and $W$ are the depth map resolution. To generate a 3D shape that is both realistic and faithful to the diversity distribution, we adopt the diffusion process \cite{sohl2015deep,ho2020denoising} that gradually denoise $N$ depth maps. These depth maps can be fused to obtain a dense point cloud, which can optionally be used to reconstruct \cite{peng2021shape,kazhdan2006poisson} a high-quality mesh model. We illustrate the entire pipeline in \cref{fig:pipeline}

In the diffusion process, we first create the ground truth multi-view depth diffusion distribution $q(\mathbf{x}_{0:T})$ in a \textit{forward process}. In this process, we gradually add Gaussian noise to each ground truth depth map $\mathbf{x}_0^v$ for $T$ steps, obtaining $N$ depth maps of pure Gaussian noise $\mathbf{x}_T = \{\mathbf{x}_T^v | v = 1,2,...,N \}$. The joint distributions can be factored into a product of per-view Markov chains:
\begin{align}
q(\mathbf{x}_{0:T}) &= q(\mathbf{x}_0)\prod^T_{t=1}{q(\mathbf{x}_t|\mathbf{x}_{t-1})} \notag \\ 
                    &= q(\mathbf{x}^{1:N}_0)\prod^N_{v=1}\prod^T_{t=1}{q(\mathbf{x}^v_t|\mathbf{x}^v_{t-1})}, \label{eq:q0T} \\
                    q(\mathbf{x}^v_t|\mathbf{x}^v_{t-1}) &:= \mathcal{N}(\mathbf{x}^v_t; \sqrt{1-\beta_t}\mathbf{x}^v_{t-1}, \beta_t\mathbf{I}), \label{eq:qtt-1}
\end{align}
where $\beta_t$ is the step $t$ variance schedule at step $t$ shared across views.

We then learn a diffusion denoising model to predict the distribution of a \textit{reverse process} $p_{\theta}(\mathbf{x}_{0:T})$ to iteratively denoise the $\mathbf{x}_T$ back to the ground truth $\mathbf{x}_0$. The joint distribution can be formulated as:
\begin{align}
    p_{\theta}(\mathbf{x}_{0:T}) &= p(\mathbf{x}_T)\prod^T_{t=1}{p_{\theta}(\mathbf{x}_{t-1}|\mathbf{x}_t)} \notag \\ 
                    &= p(\mathbf{x}^{1:N}_T)\prod^N_{v=1}\prod^T_{t=1}{p_{\theta}(\mathbf{x}^v_{t-1}|\mathbf{x}^v_t)},  \label{eq:p0T} \\
    {p_{\theta}(\mathbf{x}^v_{t-1}|\mathbf{x}^v_t)} &:= \mathcal{N}(\mathbf{x}^v_{t-1}; \mathbf{\mu}_{\theta}(\mathbf{x}^v_{t}, t), \beta_t\mathbf{I}), \label{eq:pt-1t}
\end{align}
where $\mathbf{\mu}_{\theta}(\mathbf{x}^v_{t}, t)$ estimates the mode of depth map distribution for view $v$ at step $t-1$. 

However, following \cref{eq:p0T} and \cref{eq:pt-1t},  diffusion process denoises each view independently. Starting from $N$ maps of pure random noise, a well-trained model of this kind would generate realistic depth maps $\mathbf{x}_0^{1:N}$, which however could not be fused into an intact shape due to no 3D consistency across views. 
Therefore, we propose to condition denoising steps for each view on its \NumNeighbors~neighboring views $\mathbf{x}^{r_1:r_\NumNeighbors}_t$ and replace \cref{eq:p0T} and \cref{eq:pt-1t} with:
\begin{align}
     &p_{\theta}(\mathbf{x}_{0:T})  
                    = p(\mathbf{x}^{1:N}_T)\prod^T_{t=1}\prod^N_{v=1}{p_{\theta}(\mathbf{x}^v_{t-1}|\mathbf{x}^v_{t},\mathbf{x}^{r_1:r_\NumNeighbors}_t)},  \label{eq:p0Tatt} \\
    &{p_{\theta}(\mathbf{x}^v_{t-1}|\mathbf{x}^v_{t},\mathbf{x}^{r_1:r_\NumNeighbors}_t)} := \mathcal{N}(\mathbf{x}^v_{t-1}; \mathbf{\mu}_{\theta}(\mathbf{x}^v_{t},\mathbf{x}^{r_1:r_\NumNeighbors}_t, t), \beta_t\mathbf{I}). \label{eq:pt-1tatt}
\end{align}
MVDD achieves this through an efficient epipolar `line segment' attention (\cref{sec:epipolar}). Additionally, even though the denoising process is multi-view conditioned, back-projected depth maps are still not guaranteed to be perfectly aligned in 3D space. Inspired by multi-view stereo methods \cite{galliani2015massively,schonberger2016pixelwise,xu2022point}, MVDD conducts \DepthFusion~ (\cref{sec:fusion}) in each diffusion step (\cref{eq:pt-1tatt}).



\subsubsection{Epipolar Line Segment Attention}
\label{sec:epipolar}
To promote consistency across all depth maps, we introduce an attention module named epipolar ``line segment'' attention. With the depth value of current step, MVDD leverages this information and attends only to features from visible locations on other views. To be specific, we sample on the line segment centered by the back-projected depth, rather than on the entire epipolar line \cite{shi2023mvdream,tseng2023consistent}. This design allows the proposed attention to obtain more relevant cross-view features, making it excel in both efficiency and effectiveness. The attention is defined as:
\begin{equation}
\label{eq:cross}
\begin{aligned}
Q & \in  \mathbb{R}^{(B \times N \times H \times W) \times 1 \times F}, \\
K, V & \in  \mathbb{R}^{(B\times N \times H \times W) \times (\NumNeighbors \times k) \times F}, \\
\operatorname{Cross-Attn}(Q, & K, V)= \operatorname{softmax}\left(\frac{Q K^T}{\sqrt{d_k}}\right) V,
\end{aligned}
\end{equation}
where $B$ is the batch size, $N$ is the total number of views, $k$ is the number of samples along the epipolar ``line segment'', $\NumNeighbors$ is the number of neighboring views and $F$ is the number of feature channels. At denoising step $t$, for any pixel $\SourcePixel$ at a source depth map $\mathbf{x}^v_t$, we first back project its depth value $\SourcePixelDepth$ into the 3D space to obtain a 3D point $\threeDpoint^{v_{ij}}$, and project it to a coordinate $\NeighborPixel$ on neighboring view $r$:
\begin{align}
\threeDpoint^{v_{ij}} &= ~\SourcePixelDepth A^{-1} \SourcePixel, \text{where}~\SourcePixel := [i,j,1]^T,  \label{eq:3dpoint} \\
\NeighborPixel&=A~\pi_{v \rightarrow r}\threeDpoint^{v_{ij}}  \label{eq:neighborpixel},
\end{align}
where $\threeDpoint^{v_{ij}}$ is in the camera coordinate of view $v$ , $\pi_{v \rightarrow r}$ is the relative pose transformation matrix and $A$ is the intrinsic matrix. 
Since $\SourcePixelDepth$ is noisy, we select another $k-1$ evenly spaced points around $\threeDpoint^{v_{ij}}$ along the ray and project these points, $\{\threeDpoint_1^{v_{ij}},...,\threeDpoint_k^{v_{ij}}\}$, into each neighboring view, as shown in \cref{fig:pipeline}~(right). The $k$ projected pixels lay on a epipolar ``line segment'' on view $r$ and provides features for $K, V$ in \cref{eq:cross}.

\begin{figure*}[hbt]
  \centering
   \includegraphics[width=\linewidth]{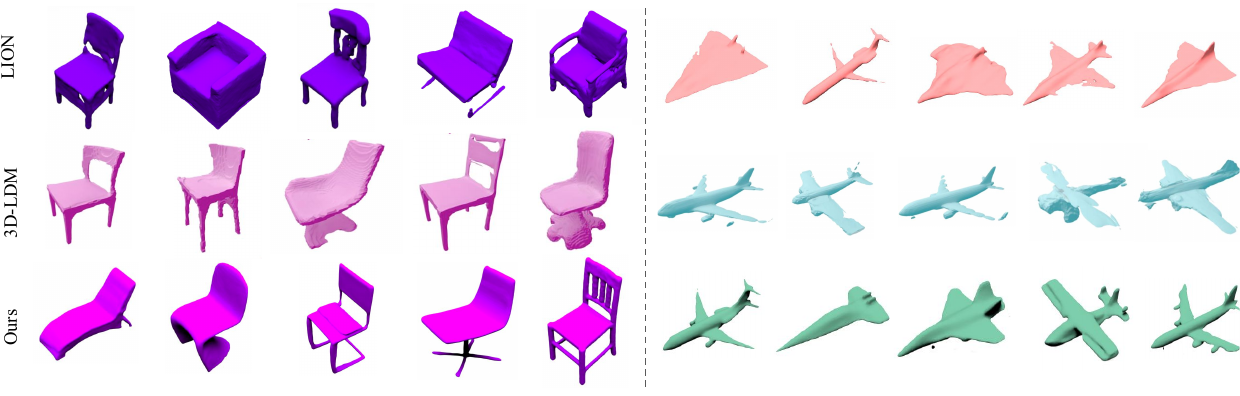}\vspace{-4mm}
    \hfill\subcaptionbox{Chair}[0.43\linewidth]
    \hfill\subcaptionbox{Airplane}[0.40\linewidth]
    
   \caption{Our generated meshes exhibit superior quality compared to point cloud diffusion model~\cite{vahdat2022lion} and implicit diffusion model~\cite{nam20223d}.}
   \label{fig:mesh}
\end{figure*}

\paragraph{Cross attention thresholding.} To ensure that depth features from a neighboring view $r$ are relevant to $\SourcePixelDepth$, we need to cull only the $\NeighborPixel$ that are also visible from source view $v$. Let $z(\cdot)$ denote the operator to extract the $z$ value from a vector $[x,y,z]$, we create the visibility mask by thresholding the Euclidean distance between the depth value of the 3D point in $r$'s camera coordinate, $\threeDpoint^{r_{mn}} = \pi_{v \rightarrow r}\threeDpoint^{v_{ij}} $, and the predicted depth value on the pixel $\NeighborPixel$ that $\threeDpoint^r$ projects onto:
\begin{equation}
M(\NeighborPixel)=\left\| z(\pi_{v \rightarrow r}\threeDpoint^{v_{ij}}) - \NeighborPixelDepth \right\|<\tau.
\end{equation}
For projected pixels that do not satisfy the above requirement, in \cref{eq:cross}, we manually overwrite their attention weights as a very small value to minimize its effect.

\paragraph{Depth concatenation.} 
For pixel $v_{ij}$, since the sampled points  $\{\threeDpoint_1^{v_{ij}},...,\threeDpoint_k^{v_{ij}}\}$ query geometric features $K,V$ from neighboring views, the attention mechanism conditions the denoising step of $\mathbf{x}^{v_{ij}}_t$ with the features $V$ weighted by the similarity between $Q$ and $K$. To enhance awareness of the locations of these points, we propose concatenating the depth values $\{z(\threeDpoint_1^{v_{ij}}),...,z(\threeDpoint_k^{v_{ij}})\}$ to the feature dimension of $V$, resulting in the last dimension of $V$ being $F+1$.

The intuition behind this is if the geometric features of $v_{ij}$ are very similar to features queried by $\threeDpoint_1^{v_{ij}}$, the depth value $\mathbf{x}^{v_{ij}}_{t-1}$ should move toward $z(\threeDpoint_1^{v_{ij}})$.
We empirically verify the effectiveness of the depth concatenation in Tab.~\ref{tab:ablation}.

\subsubsection{\DepthFusionCap~}
\label{sec:fusion} 
To further enforce alignment across multi-view depth maps, MVDD incorporates depth fusion in diffusion steps during ancestral sampling. 

Assuming we have multi-view depth maps $\{\mathbf{x}_1,..., \mathbf{x}_N\}$, following multi-view stereo methods
\cite{merrell2007real,yao2018mvsnet}, a pixel $\SourcePixel$ will be projected to another view $r$ at $\NeighborPixel$ as described in \cref{eq:3dpoint}. Subsequently, we reproject $\NeighborPixel$ with its depth value $\mathbf{x}^{\NeighborPixel}$ towards view $v$:
\begin{align}
  \threeDpoint^{\BackPixel} &= \pi_{r \rightarrow v} \mathbf{x}^{\NeighborPixel}A^{-1}\NeighborPixel, \label{eq:back3dpoint} \\
  \BackPixel &=A\threeDpoint^{\BackPixel} \label{eq:backpixel},
\end{align}
where $\threeDpoint^{\BackPixel}$ is the reprojected 3D point in view $v$'s camera coordinate. 
To determine the visibility of pixel $\SourcePixel$  from view $r$, we set two thresholds:
\begin{align}
   \left\| \SourcePixel - \BackPixel   \right\| < \PixelThreshHold , ~
   \frac{ | \mathbf{x}^{\SourcePixel} -  z(\threeDpoint^{\BackPixel}) |}{\mathbf{x}^{\SourcePixel}} < \epsilon_\theta,
\end{align}
where $z(\threeDpoint^{\BackPixel})$ represents the reprojected depth,  $\PixelThreshHold$ and $\epsilon_\theta$ are the thresholds for discrepancies between reprojected pixel coordinates and depth compared to the original ones.

\paragraph{Integration with denosing steps.} 
For a diffusion step $t$ described in \cref{eq:pt-1tatt}, after obtaining  $\mathbf{\mu}_{\theta}(\mathbf{x}_{t}, t)$, we apply \textit{depth averaging}. For each pixel, we average the reprojected depths from other visible views to refine this predicted value. Subsequently, we add $\mathcal{N}(\mathbf{0},\beta_t\mathbf{I})$ on top to obtain $\{\mathbf{x}^v_{t-1} | v = 1,2,...,N\}$. Only at the last step, we also apply \textit{depth filtering} to $X_0$ to filter out the back-projected 3D points that are not visible from neighboring views.

\subsection{Training Objectives}
\label{sec:training}

Aiming to maximize $p_{\theta}(\mathbf{x}_{0:T})$, we can minimize the objective, following DDPM~\cite{ho2020denoising}:
\begin{equation}
\begin{aligned}
L_t & =\mathbb{E}_{t \sim[1, T], \mathbf{x}_0, \boldsymbol{\epsilon}_t}\left[\left\|\boldsymbol{\epsilon}_t-\boldsymbol{\epsilon}_\theta\left(\mathbf{x}_t, t\right)\right\|^2\right] \\
& =\mathbb{E}_{t \sim[1, T], \mathbf{x}_0, \epsilon_t}\left[\left\|\boldsymbol{\epsilon}_t-\boldsymbol{\epsilon}_\theta\left(\sqrt{\bar{\alpha}_t} \mathbf{x}_0+\sqrt{1-\bar{\alpha}_t} \boldsymbol{\epsilon}_t, t\right)\right\|^2\right],
\end{aligned}
\end{equation}
where $\mathbf{x}_0$ is the ground-truth multiview depth maps, $\beta_t$ and $\bar{\alpha}_t:=\prod^t_{s=1}{(1-\beta_s)}$ are predefined coefficients of noise scheduling at step $t$. 

\subsection{Implementation Details} 
\label{sec:implementation}


Our model is implemented in PyTorch~\cite{paszke2019pytorch} and employs the Adam optimizer~\cite{kingma2014adam} with the first and the second momentum set to $0.9$ and $0.999$, respectively, and a learning rate of $2e^{-4}$ to train all our models. Unless otherwise noted, we set the height $H$ and width $W$ of depth map to both be 128 and number of views of depth map $N$ to be 8. The first camera is free to be placed anywhere on a sphere, facing the object center, and form a cube with the other 7 cameras. The number of sampled points along the epipolar line segment $k$ is 10. The threshold $\tau$ for cross attention thresholding is $0.15$. We apply denoising depth fusion only in the last 20 steps. For training, we uniformly sample time steps $t = 1, ... , T = 1000$ for all experiments with cosine scheduling~\cite{nichol2021improved}. We train our model on 8 Nvidia A100-80GB and the model usually converges within 3000 epochs. Please refer to supplemental material for more details on network architecture, camera setting, and other aspects.

\section{Application}
\subsection{3D Shape Generation}
\label{sec:generation}
\paragraph{Inference strategy.} Initialized as 2D maps of pure Gaussian noise, the multi-view depth maps can be generated by MVDD following ancestral sampling \cite{ho2020denoising}:
\begin{equation}
    \mathbf{x}_{t-1}=\frac{1}{\sqrt{\alpha_t}}\left(\mathbf{x}_t-\frac{1-\alpha_t}{\sqrt{1-\tilde{\alpha}_t}} \boldsymbol{\epsilon}_\theta\left(\mathbf{x}_t, t\right)\right)+\sqrt{\beta_t} \mathbf{\epsilon},
\end{equation}
where $\mathbf{\epsilon}$ follows a isotropic multivariate normal distribution. We iterate the above process for $T=1000$ steps, utilizing the effective epipolar ``line segment'' attention and \DepthFusion~. Finally, we back-project the multi-view depth maps to form a dense ($>20$K) 3D point cloud with fine-grained details. Optionally, high-quality meshes can be created with SAP~\cite{peng2021shape} from the dense point cloud.

\begin{figure*}[t]
  \centering
   \includegraphics[width=0.85\linewidth]{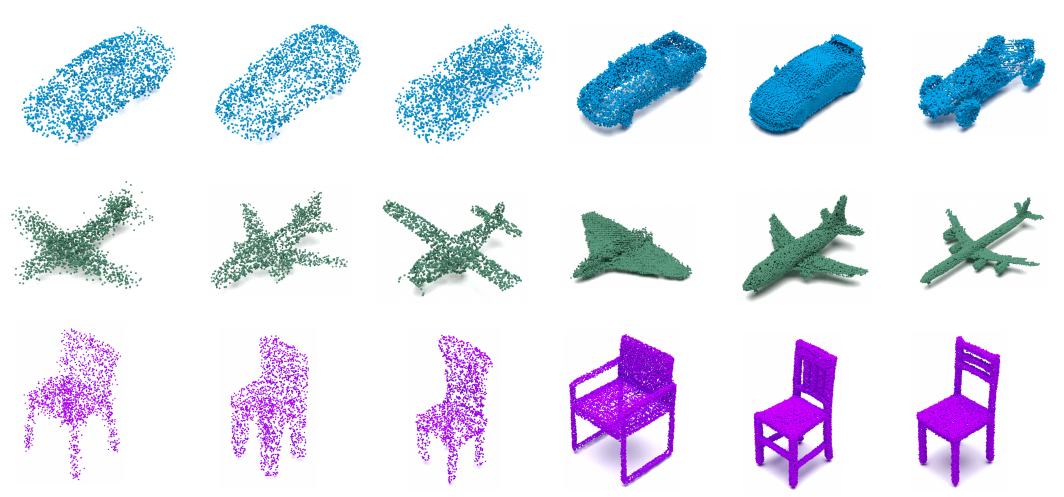}\vspace{-2mm}
   \hfill\subcaptionbox{DPM~\cite{luo2021diffusion}}[0.14\linewidth] 
    \hfill\subcaptionbox{PVD~\cite{zhou20213d}}[0.14\linewidth]
    \hfill\subcaptionbox{LION~\cite{vahdat2022lion}}[0.14\linewidth]
    \hfill\subcaptionbox{MVDD (Ours)}[0.42\linewidth]

   \caption{Unconditional generation on ShapeNet car, airplane and chair category.}
   \label{fig:uncond_generation}
\end{figure*}

\begin{table*}
\centering
 \resizebox{0.87\linewidth}{!}{
\begin{tabular}{ccccccccc}
\hline &  & Vox-diff \cite{zhou20213d} & DPM~\cite{luo2021diffusion} & 3D-LDM~\cite{nam20223d} & IM-GAN~\cite{chen2019learning} & PVD~\cite{zhou20213d} & LION~\cite{vahdat2022lion} & MVDD (Ours) \\
\hline \multirow{3}{*}{ Airplane } 
 & MMD (EMD) & 1.561 &0.990 &  3.520 & 0.980  & 1.000  & \, \, 0.920 \mycirc[gold] &  \, \, 0.920 \mycirc[gold] \\
 & $\mathrm{COV}$ (EMD) & 25.43 &  40.40 & 42.60 & 52.07  & 49.33 & 48.27 & \, \, 53.00 \mycirc[gold]\\
 & 1-NNA (EMD) & 98.13& 73.47 & 80.10& 64.04& 64.89&  63.49  & \, \, 62.50 \mycirc[gold] \\
 \hline \multirow{3}{*}{ Car } 
 & MMD (EMD) & 1.551 & 0.710 & -& 0.640  &0.820 &0.900 & \, \, 0.620 \mycirc[gold]\\
 & $\mathrm{COV}$ (EMD) & 22.15& 36.05 & - & 47.27 &39.51 & 42.59 & \, \, 49.53 \mycirc[gold]\\
 & 1-NNA (EMD) & 96.83 &  80.33 & - &  57.04 & 71.29 &65.70 & \, \, 56.80 \mycirc[gold]\\
\hline \multirow{3}{*}{ Chair } 
 & MMD (EMD) & 2.930 & 2.140 & 8.200 & 2.200 &2.330 & \, \, 1.720 \mycirc[gold] & 2.110 \\
 & $\mathrm{COV}$ (EMD) & 21.75 & 46.17 &  42.20 & 49.51 &46.47 &  50.52 & \, \, 51.55 \mycirc[gold] \\
 & 1-NNA (EMD) & 96.74 &  65.73 & 65.30 & 55.54& 56.14  &  57.31 & \, \, 54.51 \mycirc[gold] \\
\hline
\end{tabular}}
\caption{Unconditional generation on ShapeNet categories. MMD (EMD) is multiplied by $10^2$. $\mycirc[gold]$ represents the best result.}
\label{tab:uncon}
\end{table*}

\begin{figure}[t]
  \centering
   \includegraphics[width=\linewidth]{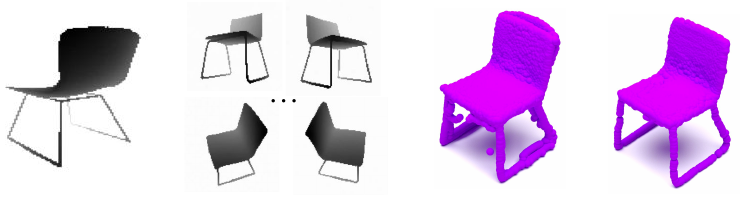}\vspace{-2mm}
    \hfill\subcaptionbox{\centering Input depth map}[0.22\linewidth] 
    \hfill\subcaptionbox{\centering Completed depth maps}[0.28\linewidth]
    \hfill\subcaptionbox{\centering W/o denoising depth fusion}[0.24\linewidth]
    \hfill\subcaptionbox{\centering W/ denoising depth fusion}[0.24\linewidth]
    
   \caption{Depth completion results prove the effectiveness of the proposed \DepthFusion~strategy (\cref{sec:fusion}).}
   \label{fig:depth_fusion}
   \vspace{-10pt}
\end{figure}

\paragraph{Datasets and comparison methods.}
To assess the performance of our method compared to established approaches, we employ the ShapeNet dataset~\cite{chang2015shapenet}, which is the commonly adopted benchmark for evaluating 3D shape generative models. In line with previous studies of 3D shape generation~\cite{zhou20213d,vahdat2022lion,yang2019pointflow,chen2019learning}, we evaluate our model on standard shape generation bench mark categories: airplanes, chairs, and cars, with the same train/test split. We compare MVDD with state-of-the-art point cloud generation methods such as DPM~\cite{luo2021diffusion}, PVD~\cite{zhou20213d} and LION~\cite{vahdat2022lion}, implicit functions-based methods such as IM-GAN~\cite{chen2019learning} and 3D-LDM~\cite{nam20223d}, as well as a voxel diffusion model Vox-diff~\cite{zhou20213d}. As our method generates varying number of points and point cloud backprojected from depth maps is not uniform, we sample 2048 points from meshes using SAP~\cite{peng2021shape} and measure against ground-truth points with inner surface removed. For those implicit methods that are not impacted by inner surface, we directly use the number reported for comparison. 

\begin{figure}[t]
  \centering
   \includegraphics[width=0.95\linewidth]{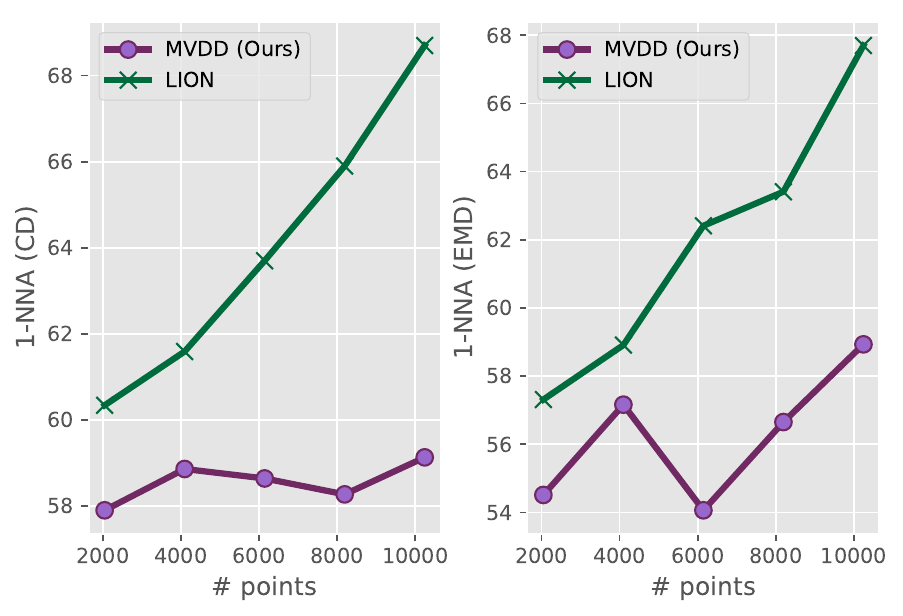}

   \caption{We report the performance of our method and LION with varying number of point clouds measured by 1-NNA  with CD and EMD, respectively, in the ShapeNet~\cite{chang2015shapenet} chair category.}
   \label{fig:upsample}
   \vspace{-10pt}
\end{figure}

\paragraph{Metrics.}  We follow \cite{zhou20213d,vahdat2022lion,chen2019learning} and primarily employ: 1) Minimum Matching Distance (MMD), which calculate the average distance between the point clouds in the reference set and their closest neighbors in the generated set; 2) Coverage (COV), which measures the number of reference point clouds that are matched to at least one generated shape; 3) 1-Nearest Neighbor Alignment (1-NNA), which measures the distributional similarity between the generated shapes and the validation set. MMD focus on the shape fidelity and quality, while COV focus on the shape diversity. 1-NNA can assess both quality and diversity of the generation results. For methods generate mesh or voxel, we transform it to point cloud and apply these metrics. Please refer to supplemental materials for more details.

\paragraph{Evaluation.}
We report the quantitative results of all methods in~\cref{tab:uncon}. Due to space constraints, we defer the performance in metric CD to the supplemental material.  Our method MVDD exhibits strong competitiveness across all categories and surpassed comparison methods, particularly excelling in the 1-NNA (EMD) metric. This metric holds significant importance as it addresses the limitations of MMD and COV~\cite{vahdat2022lion}. 

We augmented our generated point cloud and visualize the mesh quality in~\cref{fig:mesh} together with LION\cite{vahdat2022lion} and 3D-LDM~\cite{nam20223d}. Our method generates more diverse and plausible 3D shapes compared with all baselines. The visualization of our meshes shows that our method excels in synthesizing intricate details, e.g. slats of the chair and thin structure in chair base.
We also visualize point clouds in~\cref{fig:teaser}(a) and \cref{fig:uncond_generation}. The clean point cloud back-projected from our generated depth maps demonstrates 3D consistency and also validates the effectiveness of the proposed epipolar ``line segment'' attention and \DepthFusion~. In contrast,  the number of points (2048) that can be generated by point cloud-based diffusion models \cite{vahdat2022lion,zhou20213d,luo2021diffusion} limits their capabilities to capture fine-grained details of the 3D shapes.

\paragraph{Generated dense point cloud vs up-sampled sparse point cloud.}
Since our method can directly generate 20K points, while LION~\cite{vahdat2022lion} is limited to producing sparse point cloud with 2048 points, we up-sample varying number of points from LION's meshes. We then compare the performance of our method with LION. As shown in Fig.~\ref{fig:upsample}, the performance of LION deteriorates significantly as the number of points increases. It is because LION struggles to faithfully capture necessary 3D shape details with its sparse point cloud. In contrast, the performance of our method is robust with the increased number of 3D points and outperforms LION by larger margins as the point cloud density increases.


\subsection{Depth Completion}
\label{sec:dc}


\paragraph{Inference strategy.} We reuse an unconditional generative model to perform shape completion task, where depth maps from other views $\mathbf{x}^{\mathrm{other}}$ can be generated conditioned on the single input view of depth map $\mathbf{x}^{\text{in}}$. In each reverse step of the diffusion model, we have:

\begin{equation}
\label{eq:dc}
\begin{aligned}
& \mathbf{x}_{t-1}^{\mathrm{in}} ~~\sim \mathcal{N}\left(\sqrt{\bar{\alpha}_t} \mathbf{x}^\mathrm{in}_0,\left(1-\bar{\alpha}_t\right) \mathbf{I}\right), \\
\text{1st pass: ~} & \hat{\mathbf{x}}_{t-1}^{\mathrm{other}} \sim \mathcal{N}(\sqrt{1-\beta_t} ~\mathcal{\mu}_{\theta}(\mathbf{x}^{r_1:r_\NumNeighbors}_t, t), \beta_t\mathbf{I}),  \\
\text{2nd pass: ~} & \mathbf{x}_{t-1}^{\mathrm{other}} \sim \mathcal{N}(\mathcal{\mu}_{\theta}( \hat{\mathbf{x}}_{t-1}^{\mathrm{other}}, \mathbf{x}^{in}_0, t), \beta_t\mathbf{I}),
\end{aligned}
\end{equation}
where $\mathbf{x}_{t-1}^{\mathrm{in}}$ is sampled using the given depth map $\mathbf{x}^{\mathrm{in}}$, while $\mathbf{x}_{t-1}^{\mathrm{other}}$ is sampled from the model,
given the previous iteration $\mathbf{x}_t$. Different from unconditional generation, to enhance the consistency with the input view, we do two passes to denoise the other views. In the first pass each view attends to every other views and in the second pass each view only attends to the input view $\mathbf{x}^\mathrm{in}$. We scale back noise at first pass, following the Langevin dynamics steps \cite{song2020score,song2019generative}.

\begin{table}
\centering
\large
 \resizebox{\linewidth}{!}{
\begin{tabular}{ccccccccc}
\toprule & SoftFlow~\cite{kim2020softflow}  & PointFlow~\cite{yang2019pointflow} &DPF-Net~\cite{klokov2020discrete} & PVD~\cite{zhou20213d} & MVDD (Ours) \\
\hline \multirow{1}{*}{ Airplane } & 1.198 & 1.180 & 1.105 & 1.030 & \, \, 0.900 \mycirc[gold] \\
 \multirow{1}{*}{ Chair } & 3.295 & 3.649 & 3.320 & 2.939 & \, \, 2.400 \mycirc[gold] \\
 \multirow{1}{*}{Car} & 2.789 & 2.851 & 2.318 & 2.146 & \, \, 1.460 \mycirc[gold] \\
\bottomrule
\end{tabular}}
\caption{Depth completion comparison against baselines. EMD is multiplied by $10^2$. $\mycirc[gold]$ represents the best result.}
\label{tab:dc}
\end{table}

\paragraph{Datasets and comparison methods.} Following the experimental setup of PVD~\cite{zhou20213d}, we use the benchmark provided by GenRe~\cite{zhang2018learning}, which contains renderings of shapes in ShapeNet from 20 random views. For shape completion, as the ground-truth data
are involved, Chamfer Distance and Earth Mover’s Distance
suffice to evaluate the reconstruction results. We select models PointFlow~\cite{yang2019pointflow}, DPF-Net~\cite{klokov2020discrete}, SoftFlow~\cite{kim2020softflow}, and PVD~\cite{zhou20213d} for comparison.

\paragraph{Evaluation.}
We show the quantitative results of our method and baselines in~\cref{tab:dc}. Our method consistently outperforms all the baselines with EMD metric on all categories. The qualitative results in~\cref{fig:teaser}(b) also showcases that our inference strategy for depth completion can effectively ``pull" the learned depth map of other views to be geometrically consistent with the input depth map.



\subsection{3D Prior for GAN Inversion}

\begin{figure}[t]
  \centering
  \includegraphics[width=\linewidth]{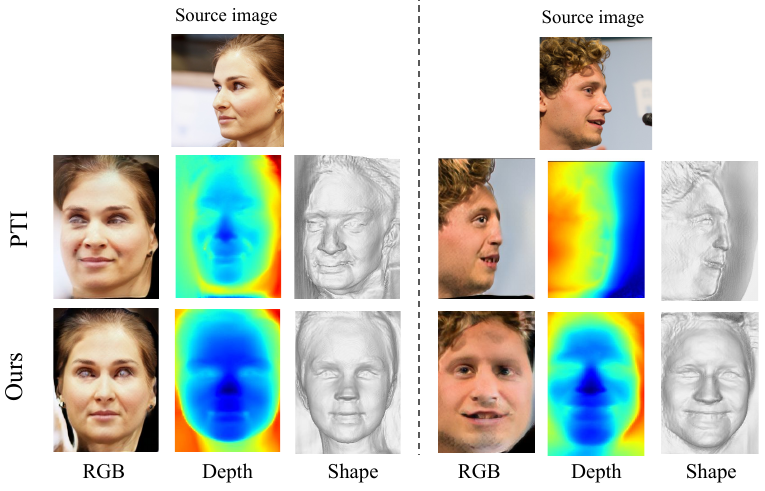}
    
  \caption{Without proper shape regularization, 3D GAN inversion~\cite{chan2022efficient} using PTI~\cite{roich2022pivotal} fails to reconstruct input image under extreme pose. Our model can serve as a shape prior for 3D GAN inversion and yield better reconstruction performance in novel frontal view.}
  \label{fig:inversion}
  \vspace{-15pt}
\end{figure}

We illustrate how our trained multi-view depth diffusion model can be plugged into downstream tasks, such as 3D GAN inversion~\cite{chan2022efficient}.
As in the case of 2D GAN inversion, the goal of 3D GAN inversion is to map an input image $I$ onto the space represented by a pre-trained unconditional 3D GAN model, denoted as $G_{\operatorname{3D}}(\cdot; \theta)$, which is defined by a set of parameters $\theta$. Upon successful inversion, $G_{\operatorname{3D}}$ has the capability to accurately recreate the input image when provided with the corresponding camera pose. One specific formulation of the 3D GAN inversion problem \cite{roich2022pivotal} can be defined as follows:
\begin{equation}
    w^*, \theta^*=\underset{w, \theta}{\arg \max }=\mathcal{L}\left(G_{3 D}(w, \pi ; \theta), I\right),
\end{equation}
where $w$ is the latent representation in $\mathcal{W}^+$ space and $\pi$ is
the corresponding camera matrix of input image. $w$ and $\theta$ are optimized alternatively, i.e., $w$ is optimized first and then $\theta$ is also optimized together with the photometric loss:
\begin{equation}
\begin{aligned}
    \mathcal{L}_{\text {photo }}=\mathcal{L}_2 & \left(G_{3 \mathrm{D}} \left(w, \pi_s ; \theta \right), I_s \right)  \\
    & +\mathcal{L}_{\mathrm{LPIPS}}\left(G_{3 \mathrm{D}}\left(w, \pi_s ; \theta\right), I_s\right),
\end{aligned}
\end{equation}
where $\mathcal{L}_{\mathrm{LPIPS}}$ is the perceptual similarity loss~\cite{zhang2018unreasonable}. 
However, with merely supervision from single or sparse views, this 3D inversion problem is ill-posed without proper regularization, so that the geometry could collapse (shown in~\cref{fig:teaser}(c) and~\cref{fig:inversion} 2nd row).
To make the inversion look plausible from other views, a 3D geometric prior is needed, as well as a pairing regularization method which can preserve diversity. Score distillation sampling has been proposed in DreamFusion~\cite{poole2022dreamfusion} to utilize a 2D diffusion model as a 2D prior to optimize the parameters of a radiance field. In our case, we use our well-trained MVDD model as a 3D prior to regularize on the multi-view depth maps extracted from the tri-plane radiance fields. As a result, the following gradient direction would not lead to collapsed geometry after inversion:
\begin{equation}
    \label{eq:inversion}
    \nabla \mathcal{L}=\nabla \mathcal{L}_{\text {photo }} + \nabla\lambda_{\mathrm{SDS}} \mathcal{L}_{\mathrm{SDS}},
\end{equation}
where $\lambda_{\mathrm{SDS}}$ is the weighting factor of $\mathcal{L}_{\mathrm{SDS}}$ \cite{poole2022dreamfusion}.

To learn the shape prior for this 3D GAN inversion task, we render multi-view depth maps from the randomly generated radiance fields of EG3D~\cite{chan2022efficient} trained with FFHQ~\cite{karras2019style} dataset. We then use them as training data and train our multi-view depth diffusion model. Using~\cref{eq:inversion}, we perform test-time optimization for each input image to obtain the optimized radiance field. In~\cref{fig:teaser}(c) and~\cref{fig:inversion}, we show the rendering and geometry results of 3D GAN inversion with and without regularization by MVDD. With the regularization of our model, the ``wall'' artifact is effectively removed and it results in better visual quality in the rendered image from novel frontal view.

\subsection{Ablation study}
\begin{table}[t]
  \begin{center}
  \resizebox{0.95\linewidth}{!}{
  \begin{tabular}{lcccccccccc} 
    \toprule
   \multicolumn{1}{c}{\centering Cross attn.} &  \multicolumn{1}{c}{\centering Depth concat.} & \multicolumn{1}{c}{\centering Cross attn. thresholding} & \multicolumn{1}{c}{\centering Depth fusion}  & \multicolumn{2}{c}{\centering 1NN-A} \\
    \cmidrule(l){5-6} 
    (\cref{sec:epipolar})  & (\cref{sec:epipolar}) &   (\cref{sec:epipolar}) & (\cref{sec:fusion}) &  CD  & EMD \\ 
    \midrule
     \XSolidBrush & \XSolidBrush  & \XSolidBrush & \XSolidBrush & 92.00 & 90.00 \\
     \Checkmark & \XSolidBrush  & \XSolidBrush  & \XSolidBrush & 61.78 & 59.65\\
     \Checkmark & \Checkmark & \XSolidBrush & \XSolidBrush & 60.72 & 59.00\\
     \Checkmark & \Checkmark & \Checkmark & \XSolidBrush & 59.82 & 57.75 \\
      \Checkmark & \Checkmark & \Checkmark &  \Checkmark &  \, \, 57.90 \mycirc[gold] & \, \, 54.51 \mycirc[gold] \\
    \bottomrule
  \end{tabular}}
 \end{center}\vspace{-3mm}
  \caption{Ablation study on the chair category. $\mycirc[gold]$ is the top result.   }
   \label{tab:ablation}
   \vspace{-15pt}
\end{table}
We perform ablation study to further examine the effectiveness of each module described in the method section. Specifically, in~\cref{tab:ablation} we report the ablated results of epipolar ``line segment'' attention, depth concatenation, and cross attention thresholding (\cref{sec:epipolar}) and depth fusion (Sec.~\ref{sec:fusion}) in ShapeNet chair category for the unconditional generation task as we describe in~\cref{sec:generation}. Without the designed cross attention, the model could barely generate plausible 3D shapes as measured by 1NN-A metric. With designs such as depth concatenation and cross attention thresholding being added, the 3D consistency along with the performance of our model is progressively improving. Last but not least, denoising depth fusion  align the depth maps and further boost the performance. Qualitatively, \cref{fig:depth_fusion} illustrates how the denoising depth fusion help eliminate double layers in depth completion task.
\section{Conclusion}
We leveraged multi-view depth representation in 3D shape  generation and proposed a novel denoising diffusion model MVDD. 
To enforce 3D consistency among different view of depth maps, we proposed an epipolar ``line segment" attention and denoising depth fusion technique. Through extensive experiments in various tasks such as shape generation, shape completion and shape regularization, we demonstrated the scalability, faithfulness and versatility of our multi-view depth diffusion model. 

{
	\small
	\bibliographystyle{ieeenat_fullname}
	\bibliography{main}
}
	

\end{document}